\documentclass[runningheads]{llncs}
\usepackage{graphicx}

\usepackage{tikz}
\usepackage{comment}
\usepackage{amsmath,amssymb} 
\usepackage{color}

\usepackage[accsupp]{axessibility}  

\usepackage{array}
\usepackage[colorlinks, linkcolor=red, anchorcolor=blue, citecolor=green]{hyperref}
\usepackage{caption}
\usepackage{subcaption}
\usepackage{caption}
\usepackage{threeparttable}
\usepackage{float}
\usepackage{makecell}
\usepackage{algorithm}
\usepackage{algpseudocode}

\begin{document}
\pagestyle{headings}
\mainmatter

\title{Boosting 3D Object Detection via Object-Focused Image Fusion} 

\titlerunning{DeMF}
%
\author{Hao Yang\thanks{Equal contribution.} \and
Chen Shi$^\star$  \and
Yihong Chen \and
Liwei Wang 
}
\authorrunning{Yang et al.}
%
\institute{Peking University \\
\email{haoy@stu.pku.edu.cn,}  \email{shichen@stu.pku.edu.cn,}\\
\email{chenyihong@pku.edu.cn,}  \email{wanglw@cis.pku.edu.cn}\\ 
}

\maketitle

\begin{abstract}

3D object detection has achieved remarkable progress by taking point clouds as the only input. However, point clouds often suffer from incomplete geometric structures and the lack of semantic information, which makes detectors hard to accurately classify detected objects. In this work, we focus on how to effectively utilize object-level information from images to boost the performance of point-based 3D detector. We present DeMF, a simple yet effective method to fuse image information into point features. 
Given a set of point features and image feature maps, DeMF adaptively aggregates image features by taking the projected 2D location of the 3D point as reference. 
We evaluate our method on the challenging SUN RGB-D dataset, improving state-of-the-art results by a large margin (+2.1 mAP@0.25 and +2.3mAP@0.5). Code is available at \href{https://github.com/haoy945/DeMF}{https://github.com/haoy945/DeMF}.

\end{abstract}


\section{Introduction}

3D object detection aims at localizing and recognizing objects in 3D scenes, which plays a vital role in various real-world applications, such as autonomous driving~\cite{wang2019monocular}, robotics manipulation~\cite{wang2019densefusion}, and augmented reality~\cite{azuma1997survey}. Among different approaches towards solving 3D object detection, methods based on point
cloud~\cite{qi2019deep,rukhovich2021fcaf3d,liu2021group,misra2021end,zhang2020h3dnet,xie2020mlcvnet,cheng2021back} have gained much attention and shown state-of-the-art performance, thanks to the depth and geometric structure information provided by LiDAR points.

\begin{figure}
\centering
\includegraphics[height=4.cm]{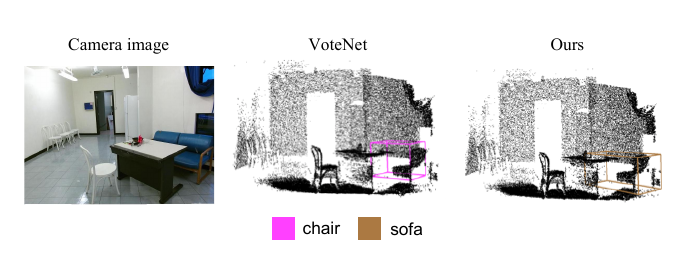}
\caption{The truncated sofa looks like a chair in point cloud. VoteNet \cite{qi2019deep} only employs information from point cloud and misclassifies the sofa as a chair. Our model can accurately recognize it by utilizing information from both point cloud and camera image.}
\label{fig:intro}
\end{figure}

While simple and efficient, incomplete geometric structure and the lack of semantic information in LiDAR points make it difficult for these point-based methods to classify objects accurately. This is demonstrated in Figure \ref{fig:intro}, where the geometric structure of the truncated sofa resembles the structure of a chair. This observation motivates us to introduce information from other modalities (e.g., RGB image) to compensate this shortage of point-based methods.

Previous works~\cite{qi2018frustum,qi2020imvotenet,10.1007/978-3-030-58555-6_3,liu2022epnet,xu2018pointfusion,Lahoud_2017_ICCV} have proposed several ways to fuse information from images and point clouds. Some works~\cite{Lahoud_2017_ICCV,xu2018pointfusion,qi2018frustum} use a pre-trained 2D detector to generate initial proposals in the form of frustums. This reduces the size of the search space of possible 3D bounding boxes, but objects missed by the 2D detector could not be recovered in the subsequent steps. Other works~\cite{qi2020imvotenet,10.1007/978-3-030-58555-6_3,liu2022epnet} take a more 3D-focused way to fuse image information into point clouds. EPNet~\cite{liu2022epnet} uses bilinear interpolation to extract image feature from the 2D projected location of the 3D point. Then the image feature is concatenated with the original point feature. ImVoteNet~\cite{qi2020imvotenet} compensates the original point feature by a feature vector containing geometric, semantic and texture cues extracted from the image. Despite getting impressive improvements over their 3D-only baselines, these methods have their limitations. In EPNet~\cite{liu2022epnet}, object-level information is missing as only the feature of the projected 2D location of the 3D point is considered. ImVoteNet~\cite{qi2020imvotenet} needs to use a trained 2D object detector to extract the category information, which makes the overall framework quite heavy. Moreover, ImVoteNet is hard to train as it needs to balance the training loss of different modalities carefully.

The defects of these methods motivate us to rethink what information we should extract from images and how they should be fused into the point features while keeping the framework succinct and efficient. To endow the point feature with the ability to classify its proposed bounding box accurately, we argue that it should be supplemented with the object-level information extracted from image. Moreover, the model should also have the ability to adaptively adjust its focus area when 
given objects of different shapes.

In this work, we solve these problems in a flexible way. Inspired by deformable convolution~\cite{dai17deform,zhu19deformv2} and deformable attention~\cite{zhu2020deformable} in the 2D object detection field, we design a novel fusion module named Deformable Attention based Multi-modal Fusion (DeMF) module to learn the sampling locations by taking the projected 2D location of the 3D point as reference. 
Extensive experiments prove that
the adaptively learned sampling strategy can effectively focus on informative locations of objects to extract object-level image features.

We validate our method on the challenging SUN RGB-D dataset~\cite{sunrgbd}. Results show that our fusion module offers significant gains over the 3D geometry only VoteNet (+5.6 mAP@0.25, +4.8
mAP@0.5), proving the usefulness of object-level information from 2D images. Moreover, based on a state-of-the-art point-based 3D detector FCAF3D that achieves 64.2 mAP@0.25 and 48.9 mAP@0.5, the proposed approach can still improve the performance by +3.2 mAP@0.25 and +2.3 mAP@0.5 respectively, reaching 67.4 mAP@0.25 and 51.2 mAP@0.5 which is the new state-of-the-art on this benchmark.

In summary, the contributions of our work are as follows:
\begin{enumerate}
    \item We conduct an in-depth analysis on what aspects can image features help 3D object detection.
    \item Based on the analysis, we propose a novel fusion module that could adaptively extract object-level image information and fuse it into LiDAR-based 3D detection pipeline.
    \item We conduct extensive ablation studies to show the effectiveness of the proposed DeMF module. It also demonstrates state-of-the-art performance in 3D object detection on the SUN RGB-D dataset. We hope our proposed method could serve as an effective baseline for the field of multi-modality 3D object detection and intrigue people to think about what is the right way to extract and fuse information from different modalities.
\end{enumerate}

\section{Related work}
\subsubsection{3D object detection based on point clouds} 

To effectively use sparse and irregular point cloud data, two different approaches are proposed: point-based and voxel-based. Powered by PointNet~\cite{qi2017pointnet} and PointNet++~\cite{qi2018pointnnetplus}, Point R-CNN \cite{shi2019pointrcnn} directly generates 3D box proposals from points, and VoteNet \cite{qi2019deep} designs a vote-based method that shifts points closer to object centers and then group points to get candidates. Some follow-up works~\cite{yang20203dssd,cheng2021back,zhang2020h3dnet,chen2020hierarchical,xie2021venet} further improve the VoteNet in speed, voting strategy and object box localization. Group-Free \cite{liu2021group} removes the voting stage and uses a transformer module instead, obtaining the feature of an object from the backbone outputs with the help of attention mechanism. 

Voxel-based methods first convert points into regular voxels and employ 3D ConvNets to process them. For example, VoxelNet~\cite{zhou2018voxelnet} uses stacked encoding layers to extract voxel features. However, voxel-based methods suffer from the large memory and computational cost when inputting large scenes. \cite{yan2018second,graham20183d} introduces a sparse convolution operation to improve the computational efficiency, which only executes convolution operations on non-empty voxels. Based on sparse 3D convolutions,~\cite{gwak2020generative,rukhovich2021fcaf3d,lang2019pointpillars,Deng2021VoxelRT,shi2020pv} can perform on par with or outperform previous methods which directly use raw point cloud data.

Notably, current 3D detection methods have achieved great success with only geometric input. To further boost detection performance, leveraging image information is a potential direction. We propose a fusion module to utilize image input, which can be integrated into these geometric-only detectors with a few modifications.

\subsubsection{3D object detection based on multiple sensors}

Obviously, semantic information contained in images is useful to 3D object detection. Early methods ~\cite{Lahoud_2017_ICCV,xu2018pointfusion,qi2018frustum,deng2017amodal} use mature 2D detectors to detect objects in 2D image, which are used to constrain the size of the 3D search space. Recent methods take a more 3D-focused way. Some works~\cite{ku2018joint,hou20193d,liang2018deep} first generate region proposals which are then enhanced by the 2D image feature. ImVoteNet \cite{qi2020imvotenet} encodes 2D detection results to guide the voting operations in VoteNet. EPNet~\cite{10.1007/978-3-030-58555-6_3} and EPNet++~\cite{liu2022epnet} fuse image features to point backbone's intermediate layers. Our fusion module shares some similarities with EPNet but differs in two
important aspects. First, the fusion operation is applied only at the end of the point backbone, so it is compatible with almost all existing detectors. Second, we use image features from multiple locations to enlarge the receptive field.

\subsubsection{Attention Mechanism in Vision}

The attention-based Transformer~\cite{vaswani2017attention} has achieved great success in the field of NLP \cite{devlin2018bert,radford2018improving}. Some works~\cite{zhu2020deformable,carion2020end,hu2018relation} also apply it into 2D object
detection. The most related work to our method is Deformable DETR \cite{zhu2020deformable}. It proposes deformable attention to capture long-range relationships and avoid unacceptable complexity. We find deformable attention is suitable for extracting image information and fusing it to points. However, direct application is not feasible and we need to make adjustments by taking the properties of 3D inputs into consideration. Specifically, we use points instead of fixed embeddings as queries, and directly use geometric mapping to get reference points. 

\section{Method}

\subsection{Revisiting deformable attention}

Deformable attention is proposed in \cite{zhu2020deformable} as a variant of multi-head attention~\cite{vaswani2017attention} to address the problem of slow convergence exists in the original DETR~\cite{carion2020end}. Different from the original multi-head attention, deformable attention only attends to a small set of nearby points around the reference point to reduce the computation cost and improve the convergence speed.

Given a query element $\boldsymbol{q}$, a 2D reference point $\boldsymbol{p}$ and an input feature map $\boldsymbol{x} \in \mathbb{R}^{C \times H \times W}$, the output of the deformable attention module can be formulated as:
\begin{equation}
    \operatorname{DeformAttn}\left(\boldsymbol{q}, \boldsymbol{p}, \boldsymbol{x}\right)=\sum_{m=1}^{M} \boldsymbol{W}_{m}\left[\sum_{k=1}^{K} A_{m k} \cdot \boldsymbol{W}_{m}^{\prime} \boldsymbol{x}\left(\boldsymbol{p}+\Delta \boldsymbol{p}_{m k}\right)\right] \label{eq:deform}
\end{equation}
where $m$ and $k$ index the attention head and the sampled key respectively. $\Delta \boldsymbol{p}_{m k} \in \mathbb{R}^{2}$ indicates sampling offset and since $\boldsymbol{p}+\Delta \boldsymbol{p}_{m k}$ is fractional, bilinear interpolation is applied when computing $\boldsymbol{x}\left(\boldsymbol{p}+\Delta \boldsymbol{p}_{m k}\right)$. $A_{m k}\in [0,1]$ denotes the attention weight of $k$-th sampling point in the $m$-th attention head, and is normalized by $\sum_{k=1}^{K} A_{m k}=1$. Here, both $\Delta \boldsymbol{p}_{m k}$ and $A_{m k}$ are calculated by linear projection over the query element $\boldsymbol{q}$.

To deal with multi-scale feature maps, \cite{zhu2020deformable} also proposes a multi-scale version of the deformable attention module, which shares a similar form to Equation \ref{eq:deform}.

\subsection{How Can Image Features Help?}

\vspace{-6pt}
\begin{figure}
\begin{subfigure}{0.47\textwidth}
    \includegraphics[width=\textwidth]{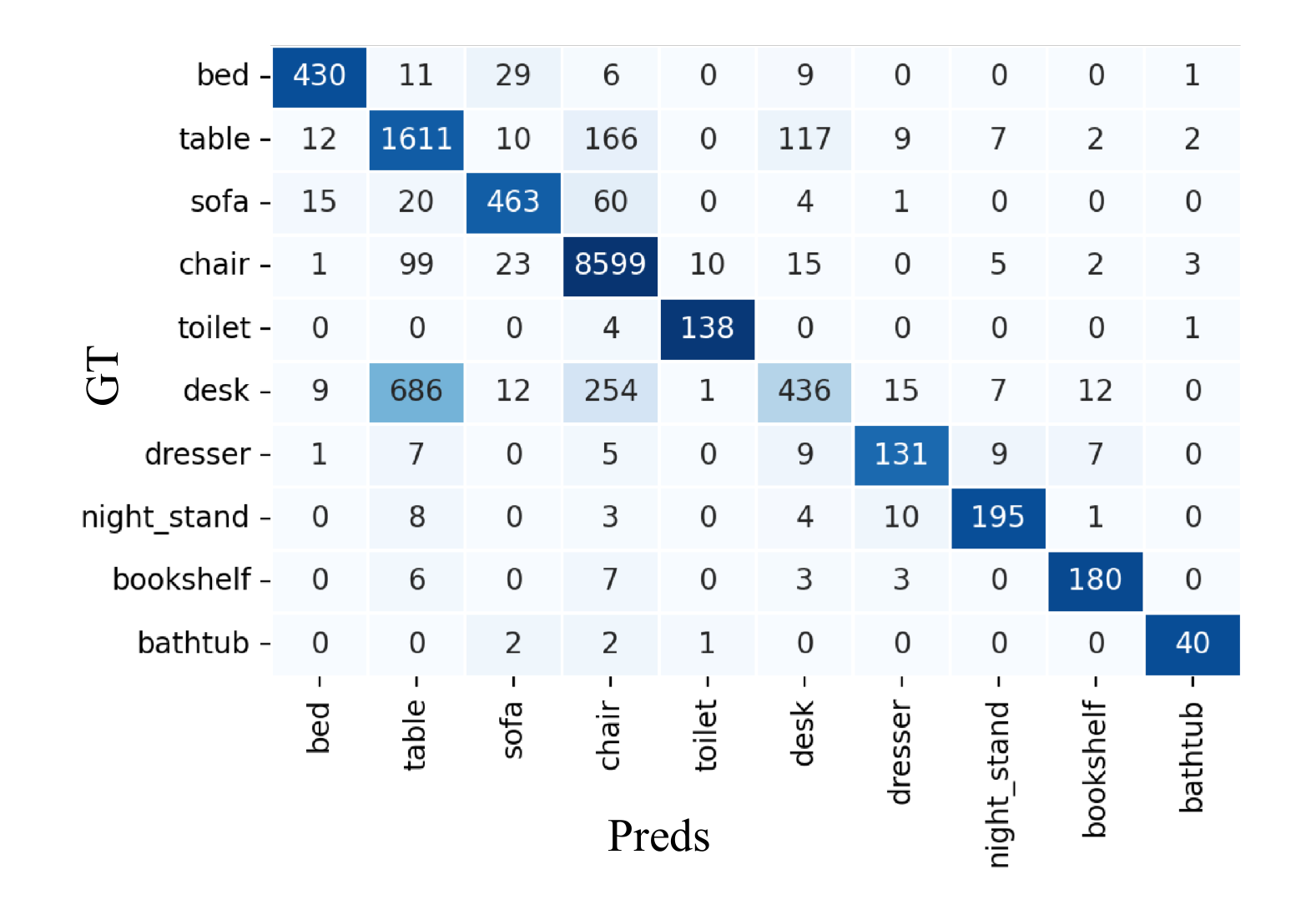}
    \caption{VoteNet}
    \label{fig:convote}
\end{subfigure}
\begin{subfigure}{0.47\textwidth}
    \centering
    \includegraphics[width=\textwidth]{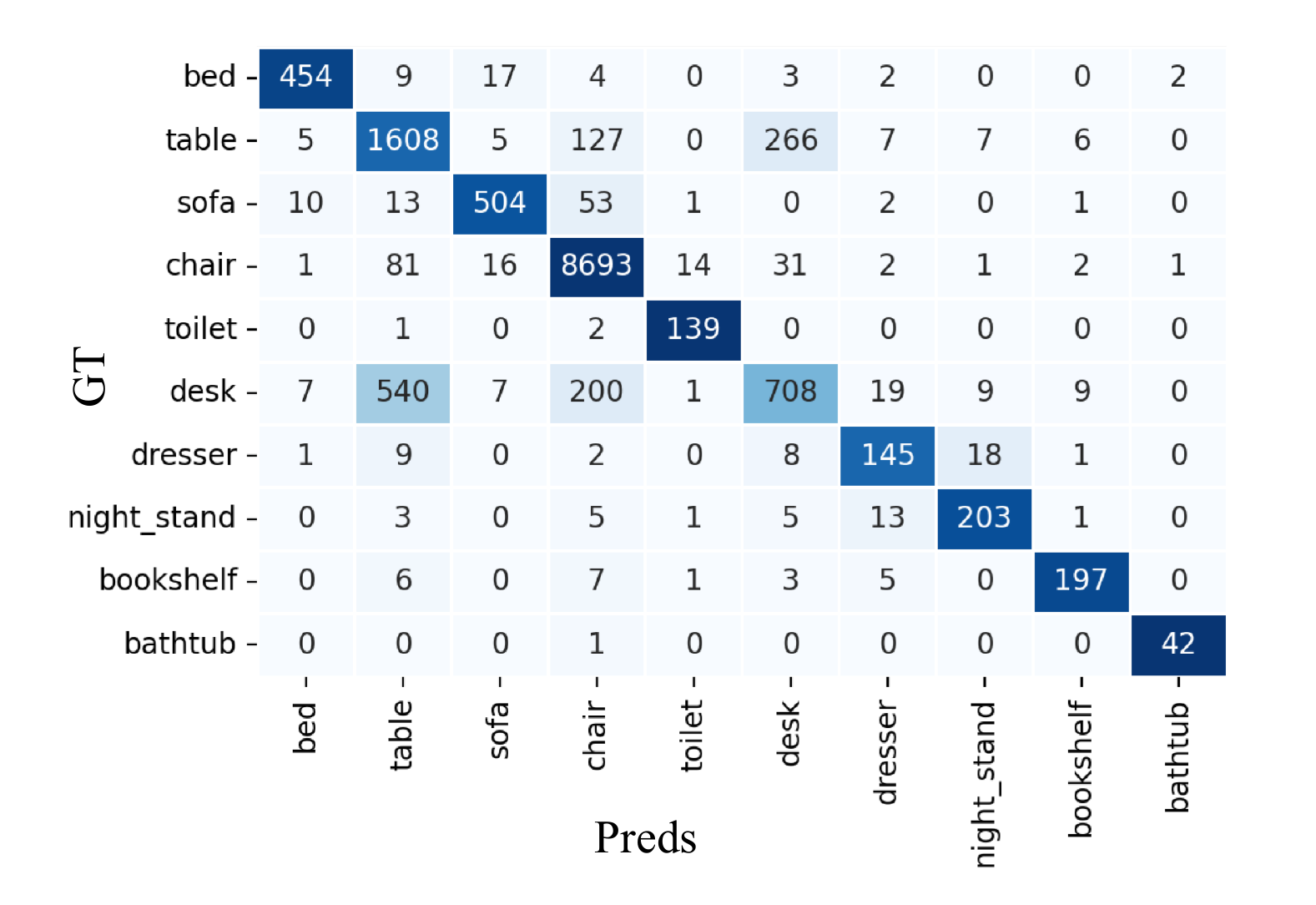}
    \caption{Ours}
    \label{fig:conour}
\end{subfigure}
\caption{Confusion matrix of VoteNet and our proposed method. Details provided in the Appendix.}
\label{fig:confusion}
\end{figure}

Given a set of $N_0$ points $\boldsymbol{S} \in \mathbb{R}^{N_0 \times 3}$ and the corresponding RGB image $\boldsymbol{x} \in \mathbb{R}^{H_0 \times W_0 \times 3}$, our goal is to boost the 3D object detection performance of point-based detection frameworks with the help of image information. To answer the question in what aspect can image features help 3D object detection, we should first investigate the defects of existing point-based detection frameworks.

LiDAR point clouds often suffer from incomplete geometric structures and the lack of semantic information, usually caused by occlusions, non-reflective surfaces, and long distances from the sensor.
As analyzed in Figure \ref{fig:intro}, VoteNet can not classify objects into the correct category in the above case.
This is further verified by Figure \ref{fig:convote}, where VoteNet struggles when classifying \textit{chair},  \textit{desk} and \textit{table}. However, it is easier to differentiate these classes when given an image. This motivates us to find a method to fuse image features into point features to enhance their recognition ability.

We argue that object-level information extracted from images could be the key to lifting the recognition performance of current point-based 3D detection methods. By fusing the object-level image feature, the corresponding point feature could sense what object it is supposed to detect. It has been proven to be effective in the field of 2D object detection that introducing object-level features into the model could help improve the detection performance~\cite{hu2018relation,dai17deform,zhu19deformv2}.

We propose an efficient fusion module named Deformable Attention based Multi-modal Fusion (DeMF) module to extract object-level information without the need for any handcrafted features. The 3D point is first projected to the 2D image plane, and then sampling locations are adaptively learned by taking the projected 2D point as reference, as shown in the right part of Figure \ref{fig:pipeline}. These learned sampling locations could focus on the semantically salient parts of the object as shown in Figure \ref{fig:refpoints}. Compared to ImVoteNet~\cite{qi2020imvotenet} which also tries to utilize object-level information, our method does not need the introduction of a trained 2D detector, which makes the whole pipeline succinct. Moreover, compared to the one-hot category information utilized in ImVoteNet, the extracted feature from the sampling points are more fine-grained and informative.

As shown in Figure \ref{fig:conour}, the proposed method could reduce the misclassification results by a lot. For example, the number of cases of classifying \textit{table} or \textit{desk} as \textit{chair} is reduced by nearly 100. 
Besides, our method can find more objects (e.g., nearly 10\% more \textit{bookshelf} are detected).
In the following section, we will elaborate the structure and training details of the proposed method.

%

\subsection{Network Structure}

\begin{figure}[t]
    \centering
    \includegraphics[height=5.4cm]{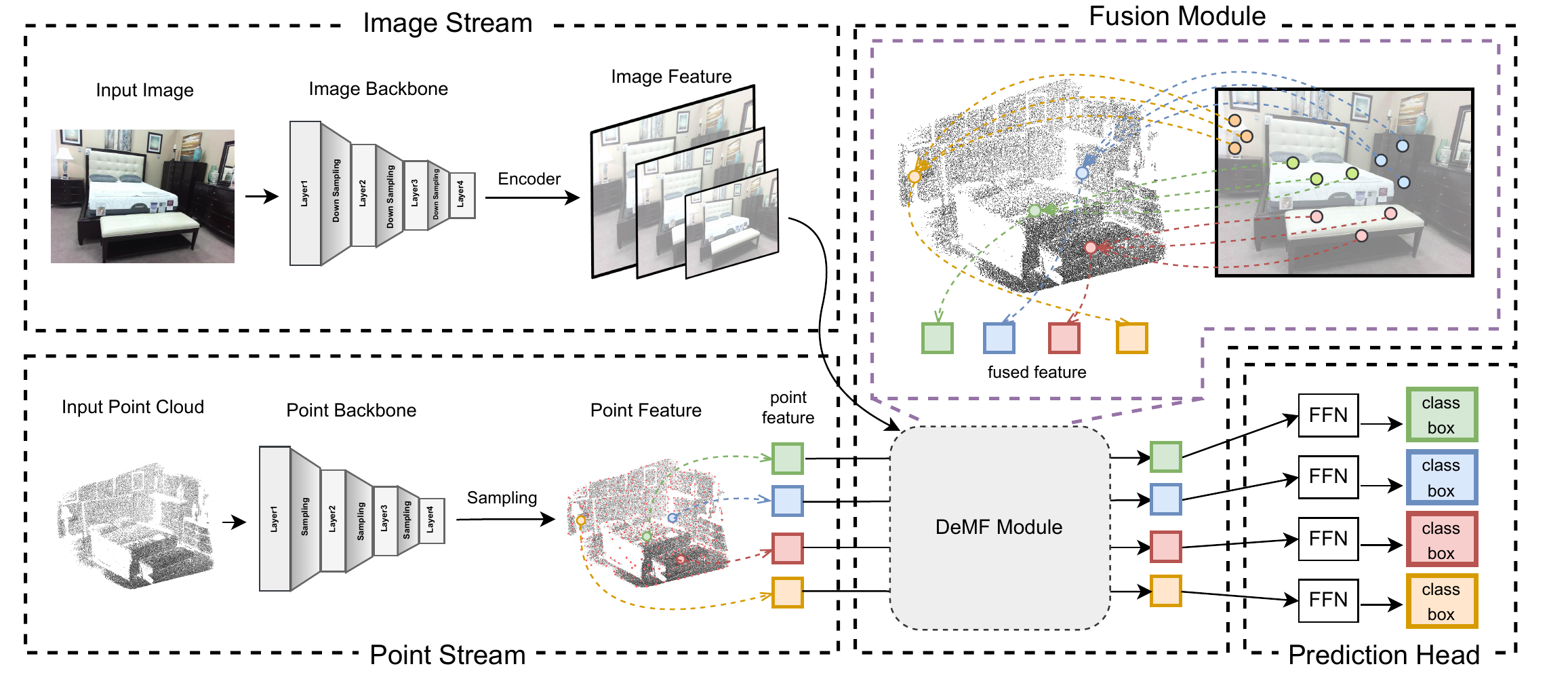}
    \caption{Illustration of the proposed pipeline. The network takes a set of points and the corresponding RGB image as input. They are processed by a point backbone and an image backbone to extract point features and image features, respectively. Then object-level image information is extracted by the proposed Deformable Attention based Multi-modal Fusion (DeMF) module and fused into point feature to enhance its recognition ability. Finally, the enhanced feature is forwarded to a feed forward network (FFN) to give the final prediction.}
    \label{fig:pipeline}
\end{figure}

 An overview of the pipeline is given in Figure \ref{fig:pipeline}. It contains three main components, (1) a two-stream backbone to extract the image and point feature representation, (2) the proposed Deformable Attention based Multi-modal Fusion (DeMF) module that aggregates image information and fuses them into point features, (3) a feed forward network (FFN) that give the final detection prediction result.

\subsubsection{Two-stream backbone}

The two-stream backbone takes point clouds and RGB images as input. Each input will be sent to the corresponding stream to get its feature representation, which the DeMF module will then process.

\paragraph{Image stream}

Given an input image $\boldsymbol{x} \in \mathbb{R}^{H_0 \times W_0 \times 3}$, a CNN backbone (e.g., ResNet), following by a deformable transformer encoder is applied to extract multi-scale feature maps $\left\{\boldsymbol{x}^{l}\right\}_{l=1}^{L} (L=4)$. Each encoder layer comprises a multi-scale deformable attention module and a feed forward network (FFN). Pixels from the multi-scale feature maps are taken as key, query and value elements. For every query pixel, the reference point is its 2D coordinates normalized by the image width and height. Every feature map is of $C = 256$ channels.

\paragraph{Point stream}

Given a set of $N_0$ points $\boldsymbol{S} \in \mathbb{R}^{N_0 \times 3}$, a point backbone (e.g., PointNet++) is applied to extract point features. Due to unacceptable complexities for processing all points, a typical point backbone usually applies a sampling module to sample high-quality object candidates (e.g., points near the object's center). Formally, a set of size $N \ll N_0$ is generated by the point stream. It can be represented as $\left\{(\boldsymbol{z}_i, \boldsymbol{s}_i)\right\}_{i=1}^{N}$, where  $\boldsymbol{z}_i$ is a $C$-channel vector representation and $\boldsymbol{s}_i$ is its corresponding 3D coordinates. The set of point features and coordinates are then used for the final bounding box prediction. Thanks to the flexibility of our proposed fusion module, there are no restrictions on the choice of the point backbone and the sampling module.

\begin{figure}[t]
    \centering
    \includegraphics[height=4.8cm]{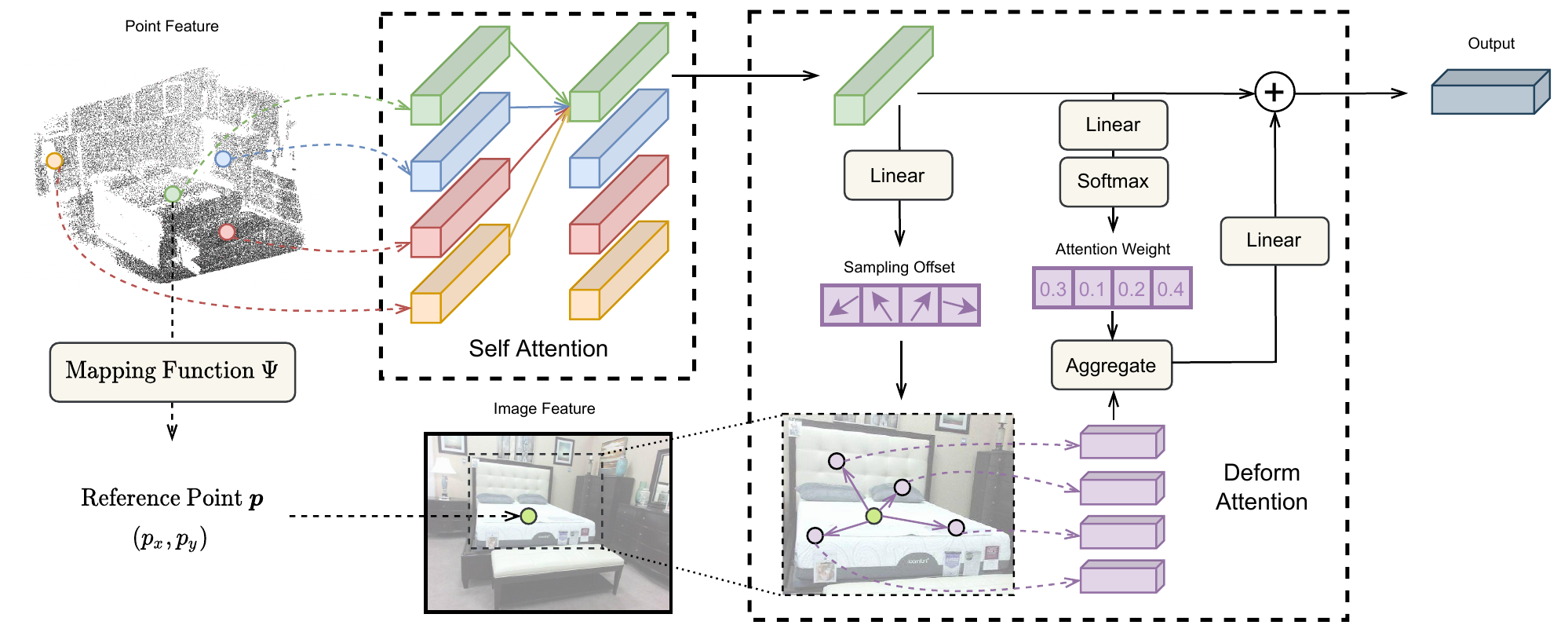}
    \caption{The architecture of the DeMF module. DeMF first uses self-attention to capture the inner relation among the input point features. Simultaneously, a mapping function $\Psi$ is applied to project the 3D coordinates to the 2D image plane as the reference points. Given the enhanced point features, the image features and the reference points as input, the deform attention module aggregates the image features extracted at the sampling locations and finally fuses them into the point features.}
    \label{fig:demf}
\end{figure}

\subsubsection{Deformable Attention based Multi-modal Fusion}

After point and image features are extracted, DeMF module is applied afterward to supplement point features with object-level information extracted from images. DeMF module is built upon the multi-scale deformable attention module, while the specific modification is applied by taking the unique properties of 3D point cloud input into consideration. The overall process of DeMF module is depicted in Figure \ref{fig:demf}.

At first, a self-attention module is applied to transfer information among the set of point features  $\left\{(\boldsymbol{z}_i, \boldsymbol{s}_i)\right\}_{i=1}^{N}$. After that, the enhanced point features, together with their corresponding 3D coordinates, interact with the multi-scale image feature maps to extract object-level information through a multi-scale deformable attention module. Through this cross-attention module, we can adaptively aggregate the image information and fuse them into point features. Finally, a feed-forward network (FFN) is applied to transform point features. This process could be stacked several times to further enhance the
feature representation. Formally it could be written as
\begin{align}
    \dot{\boldsymbol{z}}_i &= \boldsymbol{z}_i + \operatorname{SelfAttn}(\boldsymbol{z}_i, \left\{\boldsymbol{z}_j \right\}_{j=1}^{N}), \\
    \ddot{\boldsymbol{z}}_i &= \dot{\boldsymbol{z}}_i + \operatorname{MSDeformAttn}(\dot{\boldsymbol{z}}_i, \operatorname{\textit{RefPoint}}(\boldsymbol{s}_i), \left\{\boldsymbol{x}^{l}\right\}_{l=1}^{L}) \label{eq:deformattnwithrefpoint}, \\
    \hat{\boldsymbol{z}}_i &= \ddot{\boldsymbol{z}}_i + \operatorname{FFN}(\ddot{\boldsymbol{z}}_i),
\end{align}
where $i$ indexes the point feature. The two parameters of $\operatorname{SelfAttn}$ represent the query and key elements, respectively. And $\operatorname{\textit{RefPoint}}$ in Equation \ref{eq:deformattnwithrefpoint} is a mapping function that projects a 3D coordinate to a 2D coordinate, as described below.

\paragraph{Reference points}

The original deformable attention adopts a set of fixed reference points for all images, which is unsuitable for our setting, as fixed reference points may prohibit point features from obtaining corresponding image features. We expect the attention module could automatically attend to the most salient parts by taking the 3D point coordinate as prior. Inspired by this, we propose to use the mapped coordinate of 3D points in 2D images as the reference points. In this way, image features could be extracted from more relevant and informative areas.

Formally, the mapping function $\Psi: \mathbb{R}^3 \rightarrow \mathbb{R}^2$ can be denoted as:
\begin{equation}
    \Psi(x, y, z) = (\frac{\psi_1 x + \psi_2 y + \psi_3 z}{\psi_7 x + \psi_8 y + \psi_9 z}, \frac{\psi_4 x + \psi_5 y + \psi_6 z}{\psi_7 x + \psi_8 y + \psi_9 z}),
\end{equation}
where $\psi_{1 - 9}$ are the parameters of the mapping function $\Psi$ which can be calculated by the parameters of different sensors. For every point feature in the set $\left\{(\boldsymbol{z}_i, \boldsymbol{s}_i)\right\}_{i=1}^{N}$, the coordinate of reference point $\hat{\boldsymbol{p}}_i$ is calculated by
\begin{equation}
    \operatorname{\textit{RefPoint}}\left( \boldsymbol{s}_i\right)=\Phi\left(\Psi\left(s_{ix}, s_{iy}, s_{iz}\right)\right),
\end{equation}
where $\Phi$ is a normalization function that normalize the 2D coordinate into the range $[0, 1]^2$ based on the size of the image.
\begin{equation}
    \Phi(x, y) = (\frac{x}{W_0}, \frac{y}{H_0}),
\end{equation}
where $W_0, H_0$ are the width and height of the image, respectively.

\subsection{Training Details}

After point features are enhanced by image features through the DeMF module, the original prediction process and loss function $\mathcal{L}$ of the underlying point-based detector can be directly applied. 

During training, we found it is helpful to use auxiliary losses to supervise intermediate features. The same prediction process without parameter sharing and loss function are applied to each layer in the DeMF module. And the final loss is the average of the losses of all layers:
\begin{equation}
    \mathcal{L}_{\text {total}}=\frac{1}{L+1} \sum_{l=0}^{L} \mathcal{L}^{(l)},
\end{equation}
where $L$ is the number of DeMF layer. Note that we also calculate the loss on the input of the DeMF module.

Moreover, we use iterative object box prediction following \cite{liu2021group}. Box predictions from the previous layer are used to produce the refined spatial encoding via a linear layer. Then the refined spatial encoding is added to the input query of the next layer.

\section{Experiments}

\subsection{Experiments Setup}

\subsubsection{Dataset and Evaluation Protocol}

We use SUN RGB-D benchmark dataset~\cite{sunrgbd} for evaluation, which is an indoor, single-view RGB-D dataset for 3D scene understanding. The dataset is composed of $\sim$10K RGB-D images with per-point semantic labels and oriented bounding object bounding boxes for 37 object categories. Following \cite{song2016deep}, we train and evaluate the model on the 10 most common categories. In order to get relatively stable results, we run training for 5 times and test each trained model for 5 times independently for each setting. We report both the best and average performance across all 5 $\times$ 5 trials. To get the point cloud data, we use the provided camera parameters to generate it.

\subsubsection{Implementation details}

We implement our module using the MMdetection3D \cite{mmdet3d2020} framework. We mainly conduct our studies based on VoteNet~\cite{qi2019deep}, which is a point-based method. The sampling module in point stream select 256 points, and we directly use voting candidates after voting stage as sampling results in VoteNet. We also conduct experiments on FCAF3D~\cite{rukhovich2021fcaf3d}. For FCAF3D, we reserve the native prediction head and choose points according to their classification scores. ResNet50 \cite{he2016deep} is utilized as the image backbone. More details are given in Appendix.

\subsection{State-of-the-art Comparison}

In this section, we compare with previous state-of-the-arts on SUN RGB-D dataset. We first compare with previous methods that all use VoteNet as the point backbone in Table \ref{table:votenetbased}. As shown in the table, our DeMF module achieves a boost of +5.6 mAP@0.25 and +4.8 mAP@0.5 over the original VoteNet, which outperforms all previous fusion methods by a large margin. Particularly, it surpasses ImVoteNet by 1.0 on mAP@0.25 and 2.7 on mAP@0.5, demonstrating the superiority of our fusion operations. 

\setlength{\tabcolsep}{5pt}
\begin{table}[t]
\begin{center}
\caption{Comparison on SUN RGB-D with different fusion methods. All methods use VoteNet as the base point detector. For our results, the reported metric is the best result and the number within the bracket is the average result. ImVoteNet${}^\dag$ indicates using Deformable DETR as the 2D image detector instead of Faster RCNN. PC stands for point cloud.}
\vspace{6pt}
\label{table:votenetbased}
\begin{tabular}{lcccc}
\Xhline{0.8pt}\noalign{\smallskip}
Methods & Point Backbone & Input & mAP@0.25 & mAP@0.5\\
\noalign{\smallskip}
\hline
\noalign{\smallskip}
VoteNet~\cite{qi2019deep}\footnotemark[1] & PointNet++ & PC & 60.0 & 41.3\\
EPNet~\cite{10.1007/978-3-030-58555-6_3} & PointNet++ & PC+RGB & 60.9 & -\\
EPNet++~\cite{liu2022epnet} & PointNet++ & PC+RGB & 61.5 & -\\
ImVoteNet~\cite{qi2020imvotenet}\footnotemark[2] & PointNet++  & PC+RGB & 64.4 & 43.3\\
ImVoteNet${}^{\dag}$~\cite{qi2020imvotenet} & PointNet++  & PC+RGB & 64.6 & 43.4 \\
Ours (VoteNet based)& PointNet++ & PC+RGB &  $\boldsymbol{65.6}$ (65.3) &  $\boldsymbol{46.1}$ (45.4) \\
\Xhline{0.8pt}
\end{tabular}
\end{center}
\end{table}
\setlength{\tabcolsep}{1.4pt}

\footnotetext[1]{We report the results of our improved implementation instead of the official paper, which reported 57.7 mAP@0.25.}
\footnotetext[2]{We report the results of our improved implementation instead of the official paper, which reported 63.4 mAP@0.25.}

Next, we compare with previous state-of-the-art methods. The results are shown in Table \ref{table:sota}. We validate the effectiveness of our fusion module on FCAF3D. When applied to this stronger backbone, our method could still achieve considerable gains, outperforming the original FCAF3D by a large margin of 3.2 and 2.3 on mAP@0.25 and mAP@0.5, respectively. We also achieve the state-of-the-art results, 67.4 mAP@0.25 and 51.2 mAP@0.5, on this benchmark.

\setlength{\tabcolsep}{5pt}
\begin{table}[t]
\begin{center}
\caption{Comparison on SUN RGB-D with state-of-the-art methods. For Group-Free, FCAF3D and our results, the reported metric is the best result and the number within the bracket is the average result. PC stands for point cloud.}
\vspace{6pt}
\label{table:sota}
\begin{tabular}{lcccc}
\Xhline{0.8pt}\noalign{\smallskip}
Methods & Point Backbone & Input & mAP@0.25 & mAP@0.5\\
\noalign{\smallskip}
\hline
\noalign{\smallskip}
3DETR~\cite{misra2021end} & Transformers & PC & 59.1 & 32.7\\
MLCVNet~\cite{xie2020mlcvnet} & PointNet++ & PC & 59.8 & -\\
H3DNet~\cite{zhang2020h3dnet} & 4$\times$PointNet++ & PC & 60.1 & 39.0\\
BRNet~\cite{cheng2021back} & PointNet++ & PC & 61.1 & 43.7\\
Group-Free~\cite{liu2021group} & PointNet++ & PC & 63.0 (62.6) & 45.2 (44.4)\\
FCAF3D~\cite{rukhovich2021fcaf3d} & HDResNet34 & PC & 64.2 (63.8) & 48.9 (48.2)\\
PointFusion\cite{xu2018pointfusion} &  PointNet & PC+RGB & 45.4 & -\\
COG\cite{7780538} &  - & PC+RGB & 47.6 & -\\
F-PointNet~\cite{qi2018frustum} &  PointNet & PC+RGB & 54.0 & -\\
ImVoteNet~\cite{qi2020imvotenet} & PointNet++  & PC+RGB & 64.4 & 43.3\\
EPNet~\cite{10.1007/978-3-030-58555-6_3} & PointNet++ & PC+RGB & 64.6 & -\\
EPNet++~\cite{liu2022epnet} & PointNet++ & PC+RGB & 65.3 & -\\
Ours (FCAF3D based)& HDResNet34 & PC+RGB & $\boldsymbol{67.4}$ (67.1) & $\boldsymbol{51.2}$ (50.5) \\
\Xhline{0.8pt}
\end{tabular}
\end{center}
\end{table}
\setlength{\tabcolsep}{1.4pt}

Table \ref{table:per_class} shows the per-class 3D object detection results on SUN RGB-D. Equipped with the DeMF module, the model gets better results on nearly all categories and has the biggest improvements on object categories that are difficult to distinguish by geometrical appearance (+19.8 AP for dressers, +8.1 AP for bookshelves and +7.9 AP for desks).

\subsection{Ablation Study}
In this subsection, we discuss the key design choices of the proposed DeMF module. By default, we report average mAP@0.25 and mAP@0.5 of 25 trails on SUN RGB-D for all experiments. If not specified, VoteNet is chosen as the base 3D detector in the point-stream.

\subsubsection{Image information}

As shown in Table \ref{table:ImVoteNet}, the DeMF module could improve the result of VoteNet by a large margin, demonstrating the importance of image information. When only self attention module is applied, the performance only improves by a little, showing that most of the performance gain comes from the image information extracted by the deformable attention module. To show that our method could use the image information more effectively, we compare our method with ImVoteNet \cite{qi2020imvotenet}, which is also VoteNet based 3D detector. It uses 2D detection boxes to guide the point voting process. Additionally, it uses 2D object classification results and RGB color as semantic and texture information to supplement point features. As shown in Table \ref{table:ImVoteNet}, these hand-designed operations can improve the performance of VoteNet by a large margin but still fall behind our proposed method. We attribute this superiority to the adaptive learning nature of the DeMF module. It can automatically focus on salient parts of the object to extract useful information. Moreover, we do not observe additional performance gain when applying our method to ImVoteNet, as shown in the last two rows of Table~\ref{table:ImVoteNet}, indicating the feature extracted by ImVoteNet is covered by the features extracted by the DeMF module.

\setlength{\tabcolsep}{1.2pt}
\begin{table}
\renewcommand\arraystretch{1.75}  
\scriptsize
\begin{center}
\caption{Per-class  3D  object  detection  results  on  SUN RGB-D. Reported metric is average precision with IoU threshold 0.25. ``bkshelf" and ``nstand'' indicate bookshelf and nightstand respectively.}
\vspace{6pt}
\label{table:per_class}
\begin{tabular}{lcccccccccccc}
\Xhline{0.8pt}\noalign{\smallskip}
Methods&Input&bathtub&bed&bkshelf&chair&desk&dresser&nstand&sofa&table&toilet&mAP \\
\noalign{\smallskip}
\hline
\noalign{\smallskip}
VoteNet\footnotemark[1] & PC & 75.7 & 83.4 & 36.0 & 78.2 & 25.9 & 26.6 & 64.7 & 66.1 & $\boldsymbol{53.6}$ & 90.2 & 60.0 \\
w/ DeMF & PC+RGB & $\boldsymbol{79.5}$ & $\boldsymbol{87.0}$ & $\boldsymbol{44.1}$ & $\boldsymbol{80.7}$ & $\boldsymbol{33.8}$ & $\boldsymbol{46.4}$ & $\boldsymbol{66.3}$ & $\boldsymbol{72.5}$ & 52.8 & $\boldsymbol{92.7}$ & $\boldsymbol{65.6}$\\
\Xhline{0.8pt}
\end{tabular}
\end{center}
\end{table}
\setlength{\tabcolsep}{1.4pt}

\setlength{\tabcolsep}{6pt}
\begin{table}
\begin{center}
\vspace{-15pt}
\caption{Ablation analysis on importance of image information.}
\vspace{6pt}
\label{table:ImVoteNet}
\begin{tabular}{ccccc}
\Xhline{0.8pt}\noalign{\smallskip}
Detector & DeMF &mAP@0.25  & mAP@0.5 \\
\noalign{\smallskip}
\hline
\noalign{\smallskip}
VoteNet & & 60.0 & 41.3\\
+ self-attn &  & 60.8 & 41.8\\ 
ImVoteNet & & 64.6 & 43.4 \\
\hline
VoteNet  & $\checkmark$ & 65.3 & $\boldsymbol{45.4}$ \\
ImVoteNet & $\checkmark$ & $\boldsymbol{65.4}$ & 45.2 \\
\Xhline{0.8pt}
\end{tabular}
\end{center}
\end{table}
\setlength{\tabcolsep}{1.4pt}

\subsubsection{Receptive fields}

Compared to previous multi-modal detectors (e.g., EPNet~\cite{10.1007/978-3-030-58555-6_3}), DeMF has a larger receptive field on the image, enabling the capture of object-level information. To show that a larger receptive field in images is helpful for improving the recognition ability, we conduct extensive experiments on components that may affect it. Results are given in Table \ref{table:receptive-field-deform-param}, \ref{table:receptive-field-deform-sample}.

\setlength{\tabcolsep}{6pt}
\begin{table}[t]
\begin{center}
\caption{Ablation analysis on sampling location distribution. ``Scales", ``Heads" and ``Samples" indicate image feature scales, attention heads and sampling locations for each scales and heads of deformable attention in DeMF, respectively. All models are VoteNet~\cite{qi2019deep} based.}
\vspace{6pt}
\label{table:receptive-field-deform-param}
    \centering
        \begin{tabular}{ccccc}
        \Xhline{0.8pt}\noalign{\smallskip}
        \#Scales & \#Samples &\#Heads &mAP@0.25& mAP@0.5 \\
        \noalign{\smallskip}
        \hline
        \noalign{\smallskip}
         1 & 1 & 1 & 63.9 & 43.7 \\
         1 & 4 & 8 & 64.5 & 44.2 \\
         4 & 1 & 8 & 65.2 & 45.2 \\
         4 & 2 & 8 & $\boldsymbol{65.3}$ & $\boldsymbol{45.4}$ \\
         4 & 4 & 8 & 65.2 & 45.3 \\
         \Xhline{0.8pt}
        \end{tabular}
\end{center}
\end{table}
\setlength{\tabcolsep}{1.4pt}

\setlength{\tabcolsep}{6pt}
\begin{table}[t]
\begin{center}
\vspace{-20pt}
\caption{Ablation analysis on sampling strategy on 2D image. ``Grid" indicates sampling in a fixed grid. ``Offset" indicates sampling in the grid with learnable offsets.}
\vspace{6pt}
\label{table:receptive-field-deform-sample}
    \centering
        \begin{tabular}{cccc}
        \Xhline{0.8pt}\noalign{\smallskip}
        Grid & Offset &mAP@0.25  & mAP@0.5 \\
        \noalign{\smallskip}
        \hline
        \noalign{\smallskip}
        $\checkmark$&   & 64.6 & 44.8 \\
        & $\checkmark$ & $\boldsymbol{65.3}$ & $\boldsymbol{45.4}$ \\
        \Xhline{0.8pt}
        \end{tabular}
\end{center}
\end{table}
\setlength{\tabcolsep}{1.4pt}

Table \ref{table:receptive-field-deform-param} ablates how the hyperparameters in DeMF would affect the results. When only one location is sampled on a single level of the feature maps, the performance degrades a lot as shown in the 1st row, since the receptive field is small. When increasing the number of heads and the number of sampling locations per head, the receptive field is enlarged and the results are getting better as shown in the 2nd row. By utilizing multi-scale feature maps, the results can get further improvement since sampling locations in different feature levels can focus on different grained details. We find the results get saturated when further increasing the number of sampling locations. So we choose the settings in the 4th row of Table \ref{table:receptive-field-deform-param} as the default setting, which has a remarkable performance improvement when compared to sampling only one location (+1.4 mAP@0.25 and +1.7 mAP@0.5).

By adaptively learning the sample location, DeMF could automatically focus on the informative parts of the object. Table \ref{table:receptive-field-deform-sample} validates its effectiveness. By replacing it with fixed grid sampling, the performance drops by 0.7 on mAP@0.25 and 0.6 on mAP@0.5 as it could not find the most suitable sampling location for objects of different shapes.

\subsubsection{Reference Points} 

The generation of the reference points would also influence the final performance as it tells the model where to focus. We ablate two different ways to get reference points. One is the mapping function we adopt, and the other one is using point feature to predict it. The latter way could only achieve 62.7 mAP@0.25 and 42.2 mAP@0.5, which is much lower than the mapping function's 65.3 mAP@0.25 and 45.4 mAP@0.5. We suspect the reason is that it is hard for the point feature to directly predict the precise location of the object in the 2D plane.

\subsection{Qualitative Results and Discussion}

\begin{figure}[t]
\centering
\includegraphics[height=3.8cm]{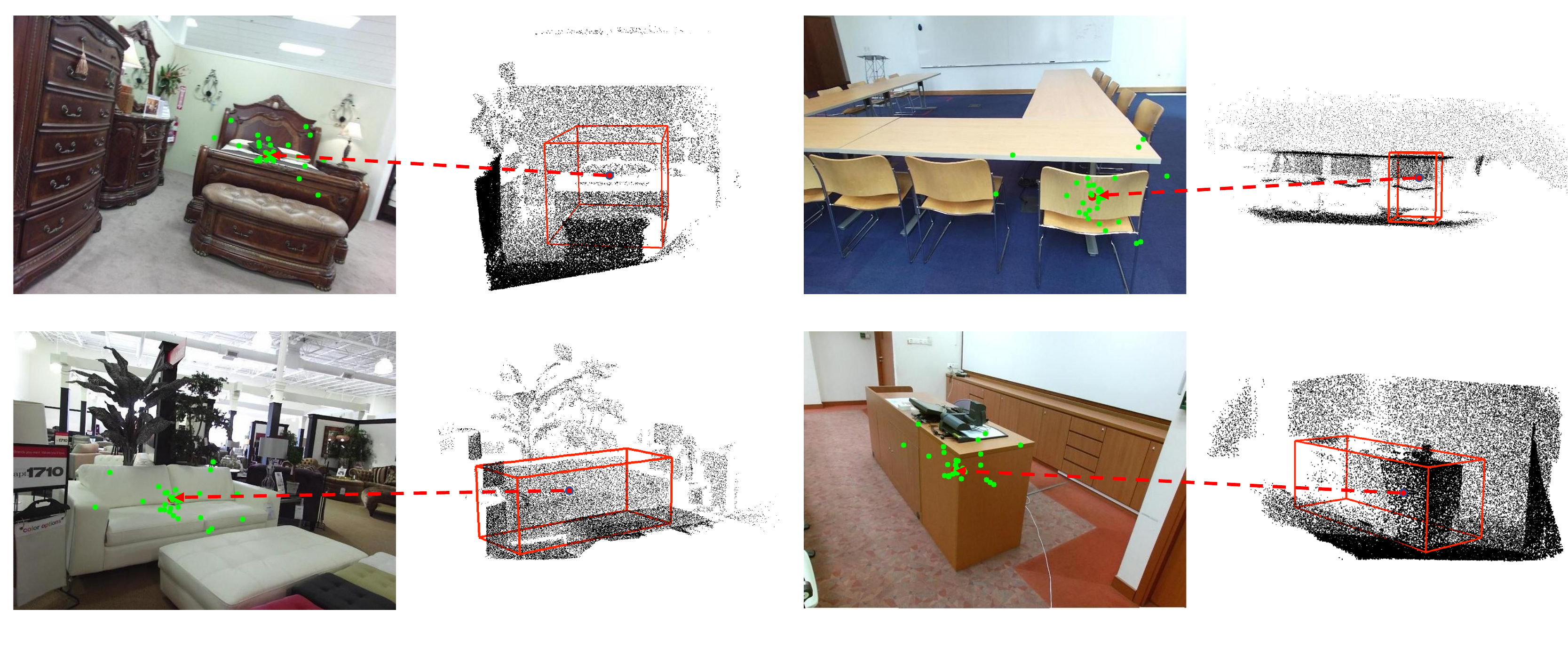}
\caption{Visualization of sample locations on the 2D image. Red dots in 2D images are reference points of red points in point clouds. Green dots in 2D images are sampling locations where image features are extracted.}
\label{fig:refpoints}
\end{figure}

In Figure~\ref{fig:refpoints}, we visualize the learned sample locations on 2D image. We can observe that some samples are clustered near the reference point and some samples locate near the boundary of the object. They together could provide meaningful object-level information for the original point feature.

\begin{figure}[t]
\centering
\includegraphics[height=6cm]{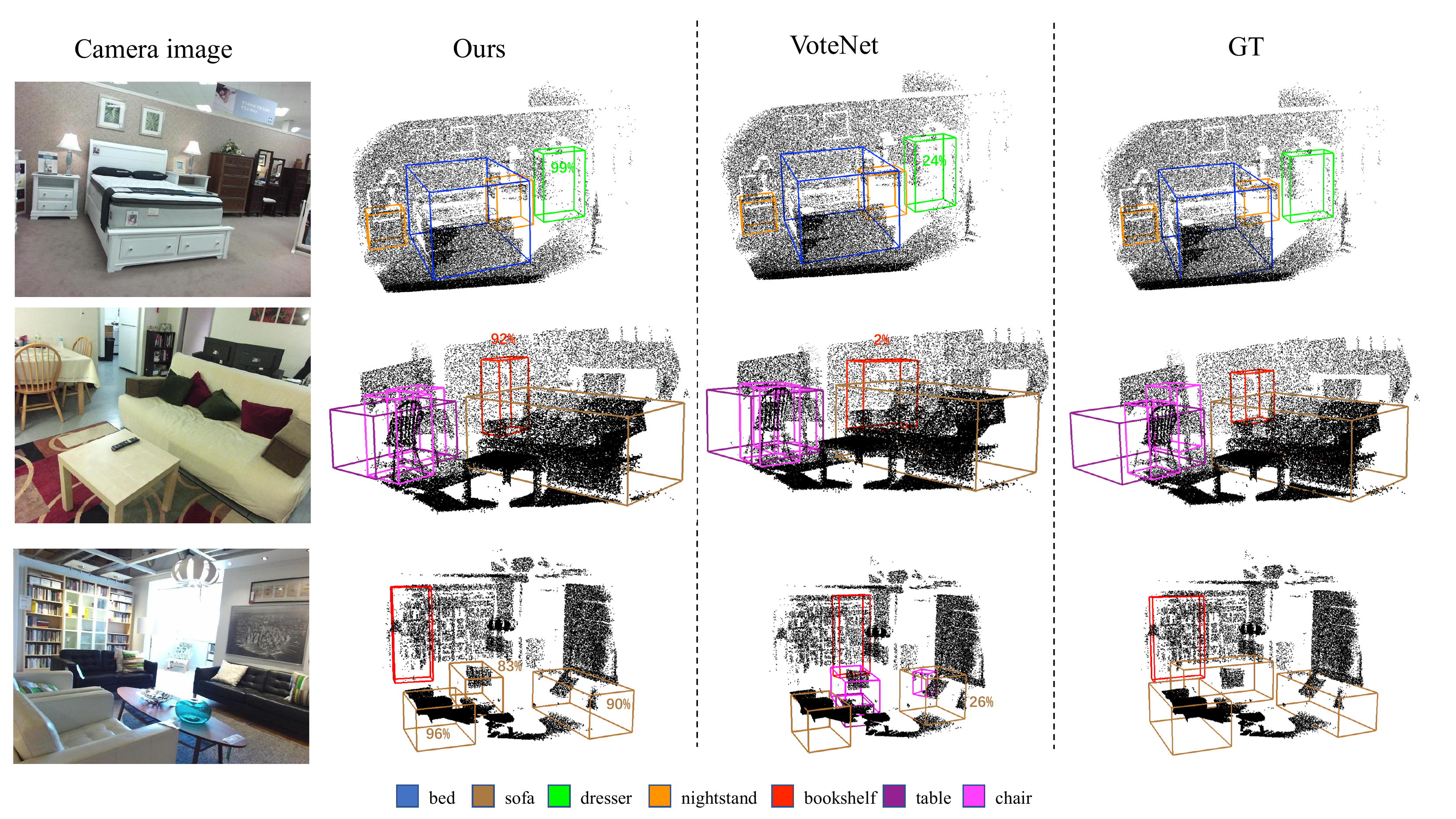}
\caption{Visualization results on SUN RGB-D Dataset. The number around the box is the detection confidence.}
\label{fig:vis_det}
\end{figure}

Figure~\ref{fig:vis_det} illustrates the qualitative results on SUN RGB-D. Detection results from the original VoteNet and VoteNet equipped with DeMF module are given. In the first example, the confidence of the detected dresser in VoteNet is only 0.24, while it is 0.99 when the DeMF module is applied. In the second example, the bookshelf is small and far away from the camera. Its geometric structure is indistinct in point cloud. Though VoteNet can find it, the classification score is only 0.02, which will be filtered in real application. However, it is easy to identify the bookshelf in the 2D image, so DeMF can help the detector to be more confident about the detection result. In the third example, there are few points located on the black sofa and this scarcity of geometric information makes VoteNet misclassify it as a chair. Thanks to the object-level information provided by the image, our model can accurately recognize it.

\section{Conclusions}

We presented DeMF, a new design for multi-modal 3D detection. 
Our method could adaptively extract object-level image information and fuse it into point features extracted by LiDAR-based 3D detection framework. 
The experimental results conducted on SUN RGB-D dataset show that our method can improve the performance of 3D detection by a large margin and achieves a new state-of-the-art.
In addition, DeMF is a flexible and general method that can be applied to most current point-based detectors.
We hope our proposed method could serve as an effective baseline for the field of multi-modality 3D object detection and intrigue people to think about what is the right way to extract and fuse information from different modalities.

\clearpage
%
%
\bibliographystyle{splncs04}
\bibliography{egbib}
\appendix
\renewcommand\thesection{\Alph{section}}

\section{Implementation Details}

\subsection{Image Stream Architecture Details}
Image stream comprises an image backbone and an image encoder. In our implementation, we use ResNet50~\cite{he2016deep} as the image backbone and deformable transformer encoder~\cite{zhu2020deformable} as the image encoder. The input multi-scale feature maps of the encoder $\left\{\boldsymbol{x}^{l}\right\}_{l=1}^{L-1} (L=4)$ are extracted from the output feature maps of stages $C_3$ through $C_5$ in ResNet50~\cite{he2016deep} (transformed by a $1\times1$ convolution). The lowest resolution feature map $x^L$ is obtained via a $3\times3$ stride 2 convolution on the final $C_5$ stage.
And every output feature map of the encoder is of $C=256$ channels.

We first train a deformable DETR~\cite{zhu2020deformable} on the training set of SUN-RGBD and then we use the pretrained parameters to initialize the image stream. Note that the parameters of the image stream are frozen when training the 3D detector.

\subsection{Point Stream Architecture Details}
We implement our method on VoteNet~\cite{qi2019deep} and FCAF3D~\cite{rukhovich2021fcaf3d} respectively. Most of the settings of point stream are the same as the original version, but with some modifications as described below.

\subsubsection{VoteNet}
We follow the setting of MMdetection3D~\cite{mmdet3d2020} implementation, but differs in three details:
(1) we use class-agnostic head for size prediction, (2) we include the IoU loss function to strengthen the supervision of the bounding box regression, (3) seed points instead of voting points are used to sample high-quality object candidates.
The above three changes can improve the performance of the VoteNet~\cite{qi2019deep}, especially in the case of mAP@0.5 (see Table~\ref{table:votenet}).

\setlength{\tabcolsep}{6pt}
\begin{table}[t]
\begin{center}
\vspace{-20pt}
\caption{Results of VoteNet~\cite{qi2019deep} for different implementation methods. ``Origin" indicates the results are from official paper. ``MMdet3D" indicates results are from MMdetection3D~\cite{mmdet3d2020} repository.}
\vspace{6pt}
\label{table:votenet}
    \centering
        \begin{tabular}{lcc}
        \Xhline{0.8pt}\noalign{\smallskip}
        Methods &mAP@0.25  & mAP@0.5 \\
        \noalign{\smallskip}
        \hline
        \noalign{\smallskip}
        Origin & 57.7 & - \\
        MMdet3D~\cite{mmdet3d2020}& 59.1 & 35.8 \\
        Ours & $\boldsymbol{60.0}$ & $\boldsymbol{41.3}$ \\
        \Xhline{0.8pt}
        \end{tabular}
\end{center}
\end{table}
\setlength{\tabcolsep}{1.4pt}

\subsubsection{FCAF3D}
The original FCAF3D~\cite{rukhovich2021fcaf3d} does not have a sampling module and will output tens of thousands of detection results, which makes it inappropriate to use the DeMF directly due to the unacceptable complexity in self-attention module. 
We add a sampling module to FCAF, which simply selects the top-$K$ ($K$=256) detection results with the highest detection scores as object candidates for DeMF.

\subsection{DeMF Architecture Details}
DeMF receives 256 point features, and each feature is of 256 channels. As for query positional embeddings, we apply linear layers on the parameterization vector of a 3D box to get the embeddings of 256 channels.

In training, the DeMF is trained from scratch, and the learning rate of DeMF module is set as 1/20 of the point stream network. 
For the VoteNet-based version, we find it helpful to ensemble the predictions of different layers in DeMF to produce final detection results, and we set the dropout rate in attention module of DeMF to 0.4.

\setlength{\tabcolsep}{1.4pt}
\begin{table}
\renewcommand\arraystretch{1.5}  
\scriptsize
\begin{center}
\caption{Per-class  3D  object  detection  results  on  SUN RGB-D. Reported metric is average precision with IoU threshold 0.25. Our results are reported by using FCAF3D as base detector. ``bkshf" and ``nstand'' indicate bookshelf and nightstand, respectively.}
\vspace{6pt}
\label{table:per_class_25}
\begin{tabular}{lcccccccccccc}
\Xhline{0.8pt}\noalign{\smallskip}
Methods&Input&bath&bed&bkshf&chair&desk&dresser&nstand&sofa&table&toilet&mAP \\
\noalign{\smallskip}
\hline
\noalign{\smallskip}
VoteNet~\cite{qi2019deep}\footnotemark[1] & PC & 75.7 & 83.4 & 36.0 & 78.2 & 25.9 & 26.6 & 64.7 & 66.1 & 53.6 & 90.2 & 60.0 \\
GroupFree~\cite{liu2021group} & PC &  $\boldsymbol{80.0}$ & 87.8 & 32.5 & 79.4 & 32.6 & 36.0 & 66.7 & 70.0 & $\boldsymbol{53.8}$ & 91.1 & 63.0 \\
FCAF3D~\cite{rukhovich2021fcaf3d} & PC & 79.0 & 88.3 & 33.0 & $\boldsymbol{81.8}$ & 34.0 & 40.1 & 71.9 & 69.7 & 53.0 & 91.3 & 64.2 \\
ImVoteNet\cite{qi2020imvotenet}\footnotemark[2] & PC+RGB & 72.8 & 87.2 & 45.1 & 80.6 & 32.7 & 38.1 & 69.0 & 72.7 & 54.9 & 91.0 & 64.4 \\
EPNet++~\cite{liu2022epnet} & PC+RGB & 76.3 & 89.1 & $\boldsymbol{47.1}$ & 80.2 & 32.5 & 45.2 & 67.4 & 71.9 & 51.3 & 92.4 & 65.3 \\
Ours & PC+RGB & 79.6 & $\boldsymbol{89.6}$ & 43.0 & 80.8 & $\boldsymbol{36.3}$ & $\boldsymbol{50.3}$ & $\boldsymbol{72.5}$ & $\boldsymbol{75.0}$ & 53.5 & $\boldsymbol{94.4}$ & $\boldsymbol{67.4}$ \\
\Xhline{0.8pt}
\end{tabular}
\end{center}
\end{table}
\setlength{\tabcolsep}{1.4pt}
\footnotetext[1]{We report the results of our improved implementation instead of the official paper, which reported 57.7 mAP@0.25.}
\footnotetext[2]{We report the results of our improved implementation instead of the official paper, which reported 63.4 mAP@0.25.}

\subsection{Training}
\subsubsection{VoteNet}
We mainly follow the training procedure in MMdetection3D~\cite{mmdet3d2020}: the network is optimized by using the AdamW optimizer ($\beta_1$=0.9, $\beta_2$=0.999) with 180 epochs. The base learning rate is 0.008 and is decreased after the 120th and 160th epochs. We train models on 8 NVidia RTX3090 with a batch size of 16.

\subsubsection{FCAF3D}
We mainly follow the training procedure in FCAF3D~\cite{rukhovich2021fcaf3d}. We use 100k points as input. The network is optimized by using the AdamW optimizer ($\beta_1$=0.9, $\beta_2$=0.999) with 36 epochs. The learning rate is set to 0.001 and decayed by 10$\times$ after 24th and 33th epochs. All the models are trained on 2 NVidia RTX3090 with a batch size of 8.

\section{Confusion matrix}
We use confusion matrix to measure the classification ability of detector. The $i$-th row in confusion matrix indicates the number of predictions assigned to gts of $i$-th category. The label assignment strategy in confusion matrix is illustrated in algorithm~\ref{alg:confusion}.

\begin{algorithm}
\caption{Label assignment in confusion matrix}
\label{alg:confusion}
\begin{algorithmic}
\State {\bfseries Input:} $S=\{s_1, s_2, ..., s_N\}$ is the list of predicted scores. $K$ and $N$ represent the numbers of ground truth boxes and predicted boxes. IOU is the iou matrix between ground truth boxes and predicted boxes, whose shape is $K \times N$.
\State {\bfseries Output:} $A=\{a_1, a_2, ..., a_N\}$, $a_i$ indicates that the predicted box $p_i$ is assigned to the $a_i$-th ground truth box.
\State {\bfseries begin}
\State \ \ \ \ Sort $S$ in descending order and get sorted index $I$
\State \ \ \ \ $U = \{False, False, ..., False\}_K$
\State \ \ \ \ {\bfseries for} $i$ {\bfseries in} $I$
\State  \ \ \ \ \ \ \ \  $iou \leftarrow IOU_{*i}$ \Comment{$IOU_{*i}$ is the i-th column of $IOU$}
\State \ \ \ \ \ \ \ \ $c \leftarrow$ argmax $iou$
\State \ \ \ \ \ \ \ \ {\bfseries if} $iou[c] \geq 0.25$ {\bfseries and} $U[i] == False$
\State \ \ \ \ \ \ \ \ \ \ \ \  $a_i = c$
\State \ \ \ \ \ \ \ \ \ \ \ \  $U[i] = True$
\State \ \ \ \ \ \ \ \ \ \ \ \  break
\State \ \ \ \ \ \ \ \ {\bfseries else}
\State \ \ \ \ \ \ \ \ \ \ \ \  $a_i = 0$ \Comment{$a_i=0$ indicates the prediction is assigned to background}
\State \ \ \ \ \ \ \ \ {\bfseries end if}
\State \ \ \ \ {\bfseries end for}
\State \ \ \ \ {\bfseries return} $A$
\State {\bfseries end}

\end{algorithmic}
\end{algorithm}

\section{Additional Results}
We present per-category results on SUN RGB-D. Table \ref{table:per_class_25} and Table \ref{table:per_class_50} show the results of mAP@0.25 and mAP@0.5, respectively. With FCAF3D, our approach gets better results on nearly all categories compared to previous methods.

\begin{figure}[H]
\centering
\includegraphics[height=7.2cm]{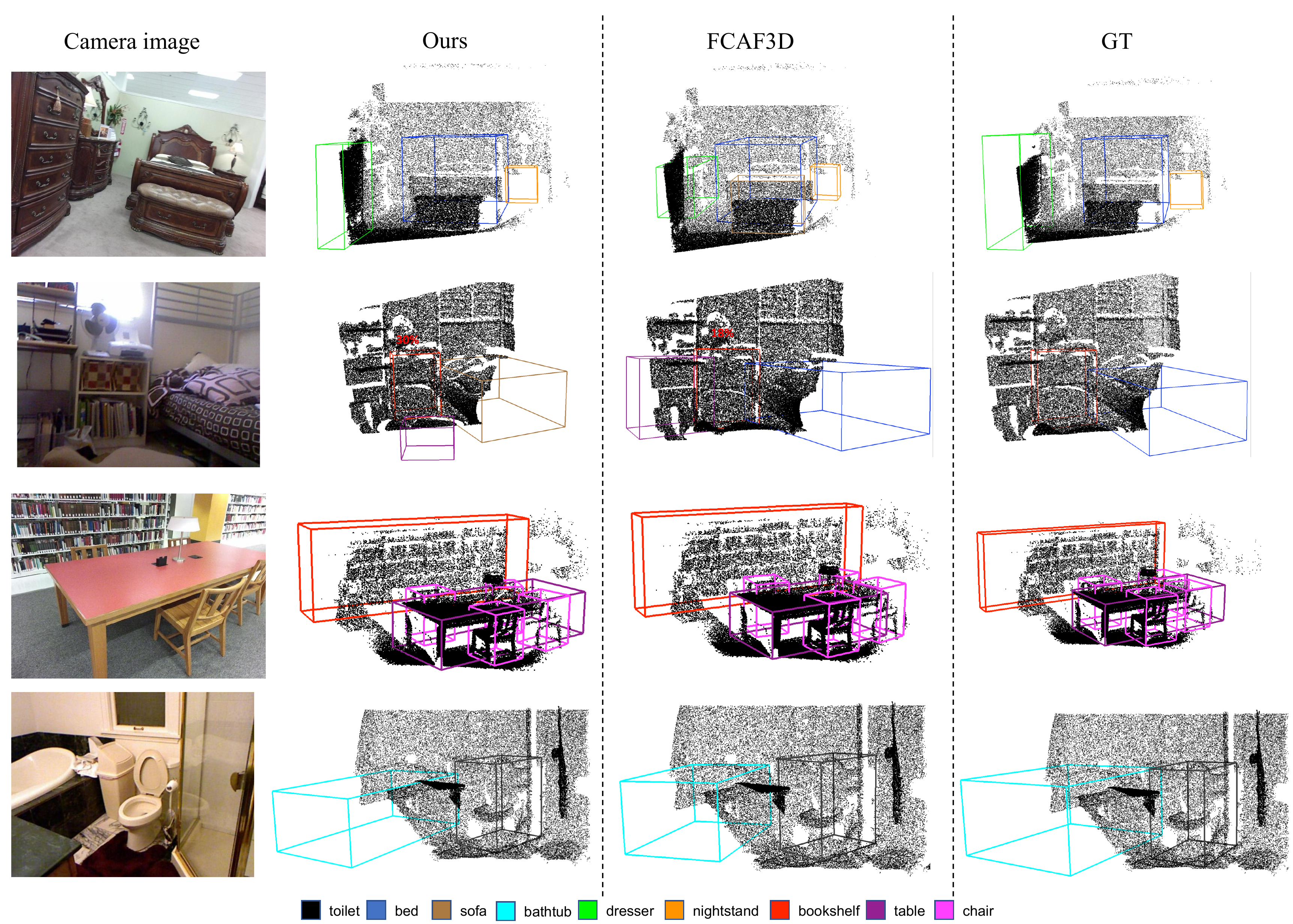}
\caption{Visualization results on SUN RGB-D Dataset.}
\label{fig:vis_fcaf3d}
\end{figure}

We provide some qualitative comparisons between original FCAF3D and FCAF3D equipped with our DeMF module on SUN RGB-D, as shown in Fig \ref{fig:vis_fcaf3d}.

\setlength{\tabcolsep}{1.4pt}
\begin{table}
\renewcommand\arraystretch{1.5}  
\scriptsize
\begin{center}
\caption{Per-class  3D  object  detection  results  on  SUN RGB-D. Reported metric is average precision with IoU threshold 0.50. Our results are reported by using FCAF3D as base detector. ``bkshf" and ``nstand'' indicate bookshelf and nightstand respectively.}
\vspace{6pt}
\label{table:per_class_50}
\begin{tabular}{lcccccccccccc}
\Xhline{0.8pt}\noalign{\smallskip}
Methods&Input&bath&bed&bkshf&chair&desk&dresser&nstand&sofa&table&toilet&mAP \\
\noalign{\smallskip}
\hline
\noalign{\smallskip}
VoteNet~\cite{qi2019deep}\footnotemark[1] & PC & 54.9 & 64.1 & 11.0 & 61.3 & 8.2 & 15.9 & 46.7 & 55.5 & 26.9 & 68.9 & 41.3 \\
GroupFree~\cite{liu2021group} & PC & 64.0 & 67.1 & 12.4 & 62.6 & 14.5 & 21.9 & 49.8 & 58.2 & 29.2 & 72.2 & 45.2 \\
FCAF3D~\cite{rukhovich2021fcaf3d} & PC & $\boldsymbol{66.2}$ & 69.8 & 11.6 & $\boldsymbol{68.8}$ & 14.8 & 30.1 & 59.8 & 58.2 & $\boldsymbol{35.5}$ & 74.5 & 48.9 \\
ImVoteNet\cite{qi2020imvotenet}\footnotemark[2] & PC+RGB & 49.6 & 60.5 & 12.9 & 63.4 & 10.5 & 20.4 & 55.4 & 59.2 & 28.9 & 72.4 & 43.3 \\
Ours & PC+RGB & 65.3 & $\boldsymbol{72.1}$ & $\boldsymbol{13.2}$ & 67.8 & $\boldsymbol{16.2}$ & $\boldsymbol{38.7}$ & $\boldsymbol{61.7}$ & $\boldsymbol{62.5}$ & 35.4 & $\boldsymbol{78.5}$ & $\boldsymbol{51.2}$\\
\Xhline{0.8pt}
\end{tabular}
\end{center}
\end{table}
\setlength{\tabcolsep}{1.4pt}
\footnotetext[1]{We report the results of our improved implementation instead of the official paper, which reported 57.7 mAP@0.25.}
\footnotetext[2]{We report the results of our improved implementation instead of the official paper, which reported 63.4 mAP@0.25.}

\end{document}